%% file: root.tex
\documentclass[letterpaper, 10 pt, conference]{ieeeconf}  

\IEEEoverridecommandlockouts                              

\usepackage{graphics} 
\usepackage{epsfig} 
\usepackage{mathptmx} 
\usepackage{times} 
\usepackage{cite}
\usepackage{amsmath,amssymb,amsfonts}
\usepackage{graphicx}
\usepackage{array}
\usepackage{textcomp}
\usepackage[left=1.4cm, right=1.4cm, top=1.7cm]{geometry}
\usepackage{caption}
\usepackage{anyfontsize}
\usepackage{subcaption}
\usepackage{hyperref}
\usepackage{pifont}
\usepackage{xurl}
\usepackage{tabularx}
\usepackage{algorithm}
\usepackage{algpseudocode}
\usepackage{algorithmicx}
\usepackage{tabularx}
\usepackage{float}
\usepackage{booktabs}

\newcolumntype{C}{>{\centering\arraybackslash}X}

\newcolumntype{M}[1]{>{\centering\arraybackslash}m{#1}}  
\newtheorem{definition}{Definition}
\algnewcommand\algorithmicforeach{\textbf{for each}}
\algdef{S}[FOR]{ForEach}[1]{\algorithmicforeach\ #1\ \algorithmicdo}

\title{\LARGE \bf
City-Wide Low-Altitude Urban Air Mobility: A Scalable Global Path Planning Approach via Risk-Aware Multi-Scale Cell Decomposition
}

\author{Josue N. Rivera$^{1}$, Dengfeng Sun$^{2}$, and Chen Lv$^{1}$
\thanks{This work was supported in part by the Ministry of Education (MOE), Singapore, under the Tier 2 Grant (MOE-T2EP50222-0002), and the NTUitive Gap Fund, Nanyang Technological University.}
\thanks{$^{1}$Josue N. Rivera and Chen Lv are with the School of Mechanical and Aerospace Engineering, Nanyang Technological University, Singapore
        {\tt\small \{josue.rivera, lyu\}@ntu.edu.sg}}%
\thanks{$^{2}$Dengfeng Sun is with the School of Aeronautics and Astronautics, Purdue University, West Lafayette, IN 47907, USA
        {\tt\small dsun@purdue.edu}}%
}

\begin{document}

\maketitle
\thispagestyle{empty}
\pagestyle{empty}

\begin{abstract}
The realization of Urban Air Mobility (UAM) necessitates scalable global path planning algorithms capable of ensuring safe navigation within complex urban environments. This paper proposes a multi-scale risk-aware cell decomposition method that efficiently partitions city-scale airspace into variable-granularity sectors, assigning each cell an analytically estimated risk value based on obstacle proximity and expected risk. Unlike uniform grid approaches or sampling-based methods, our approach dynamically balances resolution with computational speed by bounding cell risk via Mahalanobis distance projections, eliminating exhaustive field sampling. Comparative experiments against classical A*, Artificial Potential Fields (APF), and Informed RRT* across five diverse urban topologies demonstrate that our method generates safer paths with lower cumulative risk while reducing computation time by orders of magnitude. The proposed framework, Larp Path Planner, is open-sourced and supports any map provider via its modified GeoJSON 
internal representation, with experiments conducted using 
OpenStreetMap data to facilitate reproducible research in 
city-wide aerial navigation.
\end{abstract}

\vspace{4pt}
\noindent\textbf{Keywords:} Urban Air Mobility, path planning, cell decomposition, risk-aware navigation, OpenStreetMap

\input{chapters/introduction}

\input{chapters/preliminaries}

\input{chapters/larp}

\input{chapters/results}

\input{chapters/conclusion}

\addtolength{\textheight}{-12cm}   

\bibliographystyle{IEEEtran}
\bibliography{root}

\end{document}

%% file: chapters/introduction.tex
\section{Introduction}
\label{sec:introduction}

The realization of Urban Air Mobility (UAM) promises a transformative shift in metropolitan transportation, envisioning a three-dimensional ecosystem for the efficient movement of passengers and goods \cite{li2024urban, bauranov2021designing}. Within this domain, small Autonomous Aerial Vehicles (AAVs) for logistical operations such as aerial cargo delivery are anticipated to be among the earliest adopters \cite{sah2021analysis}. However, AAV deployment in dense, low-altitude urban environments introduces profound safety and scalability challenges. Unlike high-altitude airspace, the urban canyon is cluttered with static obstacles, dynamic uncertainties, and stringent regulatory restrictions, necessitating navigation solutions that prioritize operational safety and risk minimization \cite{shrestha2021survey, telli2023comprehensive, charnsethikul2025urban}.

Global path planning remains a fundamental component of autonomous navigation. Existing literature categorizes planning strategies into sampling-based and node-based approaches \cite{ghambari2024uav, mazaheri2024survey}. Sampling-based methods such as Informed-RRT* \cite{gammell2014informed, gammell2015batch} excel in high-dimensional spaces but incur high computational costs unsuitable for real-time city-scale planning. Node-based approaches like A* provide optimality and speed but suffer from memory scalability issues on high-resolution uniform grids \cite{chen2020improved, ju2020path}. Hierarchical decomposition structures (including quadtree and octree representations) reduce this combinatorial cost, yet existing implementations treat decomposition as a purely geometric operation, assigning no risk information to individual cells \cite{mazaheri2024survey}. Artificial Potential Field (APF) methods offer reactive efficiency but are prone to local minima in complex obstacle configurations \cite{hwang1992potential, rostami2019obstacle, fan2020improved}. Recent learning-based planners have also been explored for UAV navigation, though they typically require environment-specific training and lack formal safety guarantees\cite{wang2024deep}.

Recent research has pivoted toward cost-map-based planning, penalizing spatial regions based on obstacle proximity, population exposure, or ground impact risk \cite{pang2022uav, pilko2023spatiotemporal}. While effective, these methods share a common limitation: the underlying spatial representation remains a uniform discrete grid whose resolution must be set globally to guarantee safety in the narrowest passages, causing combinatorial explosion at city scale.

The core challenge is therefore finding safe paths at city scale in near-real-time, a combination no existing method fully achieves. This paper addresses this through a principled multi-scale cell decomposition in which each cell is assigned an analytically estimated risk value derived from a standardized urban risk field. Given any city's map (loaded directly from OpenStreetMap or via GeoJSON), the framework automatically computes a hierarchical decomposition where every cell carries a conservative risk bound, computed via Mahalanobis distance projection without exhaustive field sampling. This risk-annotated representation is then used by a modified A* planner that penalizes high-risk cells, naturally steering routes through safer corridors. Adaptive granularity concentrates resolution near hazards and coarsens over open airspace, making the approach both distance-competitive and scalable. Prior hierarchical decomposition methods treat subdivision as a purely geometric operation with no risk information per cell \cite{mazaheri2024survey}; cost-map approaches encode risk but over uniform grids that cause combinatorial explosion at city scale \cite{pang2022uav, pilko2023spatiotemporal}. To the authors' knowledge, Larp is the first aerial framework to assign analytically bounded risk values to individual cells of an adaptive decomposition, combining the scalability of quadtree structures with the safety guarantees of risk-field-based routing for city-scale global path planning.

We introduce \textit{Larp Path Planner} (Last-mile Autonomous Route Planning 
Path Planner), a domain-agnostic global path planning framework for restrictive 
routing within standardized ``risk fields'' \cite{rivera2024air}, demonstrated 
here through aerial cargo transport scenarios. The remainder of this paper is 
organized as follows. Section~\ref{sec:prelim} defines the standardized risk 
field. Section~\ref{sec:method} presents the multi-scale decomposition and 
risk-aware graph search. Section~\ref{sec:experiments} reports comparative 
results across five urban environments. Section~\ref{sec:conclusion} concludes 
with directions for future work.

%% file: chapters/preliminaries.tex
\section{Preliminaries}
\label{sec:prelim}
\subsection{Risk Fields}

In \cite{rivera2024air}, an unmanned aerial traffic management (UTM) system was proposed for last-mile urban air mobility that restricts UAVs to operating altitude ranges, or channels. As part of the system, a standard was introduced for defining an urban risk/cost field based on the concept of repulsion potential fields to limit routes. Physical restrictions at or above the operating altitude range of UAVs and virtual restrictions are designated as areas of high potential that diminish away from the obstacles; a behavior dictated by the individual repulsion matrix.

The standardization facilitates a uniform approach to continuous risk field construction for UAM, enabling consistent replication and analysis across different urban settings. By leveraging the structured nature of the risk field, algorithms can be developed to navigate through complex environments utilizing a common standard, avoiding obstacles, and minimizing the risk of restriction violations. The field delineated herein is henceforth denoted as a `risk field`.

\subsection{Standardized Field Units}
\label{sec:units}

Under the proposed standard, a risk field is comprised of multiple building blocks, each corresponding to a distinct type of restriction. Upon examining their definitions, a set of intrinsic properties emerges: a repulsion vector $\bar{x}$, a squared Mahalanobis distance $\tilde{d}\,^2(x)$, and a risk $\sigma(x)$. The repulsion vector $\bar{x}$ encodes the direction and proximity from the restricted areas. The squared Mahalanobis distance $\tilde{d}\,^2(x)$ reflects the weighted proximity to a point $x$, modulated by the unit's repulsion matrix $A$. Finally, the risk potential $\sigma(x)$ denotes the field's influence at point $x$, effectively reflecting the risk for restriction violation.

\begin{figure}[tb]
  \centering
    \includegraphics[width=0.25\textwidth]{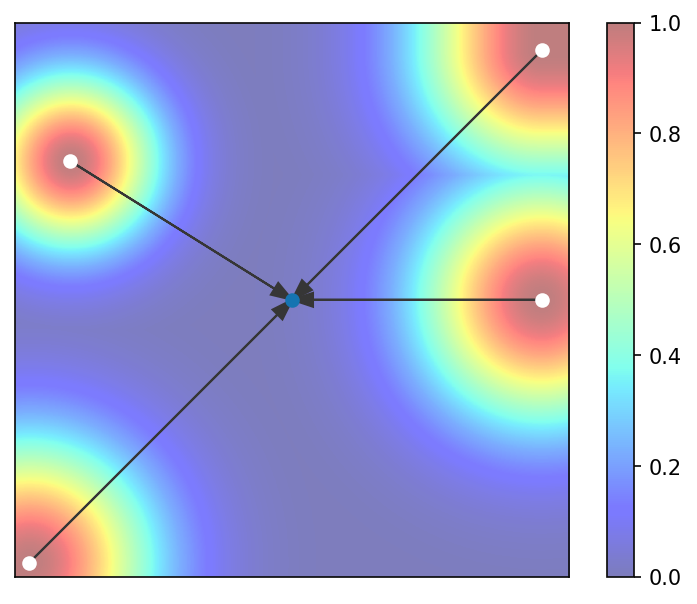}
    \caption{Repulsion vectors encoding distance and direction from nearby obstacles to an evaluation point (blue dot).}
    \label{fig:field-vectors}
\end{figure}

The repulsion vectors of the fundamental units and collection of them are detailed in Table \ref{tb:fields}. For full comparability with GeoJSON, an additional Polygon unit is included. Fig. \ref{fig:field-vectors} illustrates the repulsion vectors for an arbitrary set of obstacles with respect to an arbitrary point. The magnitude of each vector indicates proximity to the obstacle and the direction provides a guide towards moving away from it.

\begin{table*}[tbhp]
    \caption{Risk Field Unit Definitions and Repulsion Vector Formulations.}
    \label{tb:fields}
    \centering
    \begin{tabular}{|l||l|M{5.7cm}|M{2.5cm}|}
        \hline
        Unit  & Parameters & Repulsion vector $\bar{x}(x)$ & Field \\
        \hline
        Point & \begin{tabular}{@{}l@{}} Location $\hat{x}$ \\ Repulsion matrix $\mathrm{A}$ \end{tabular} & $\bar{x}_p(x) = x - \hat{x}$ & \includegraphics[width=0.09\textwidth,trim={2.5cm 1.5cm 2.25cm 2.5cm},clip]{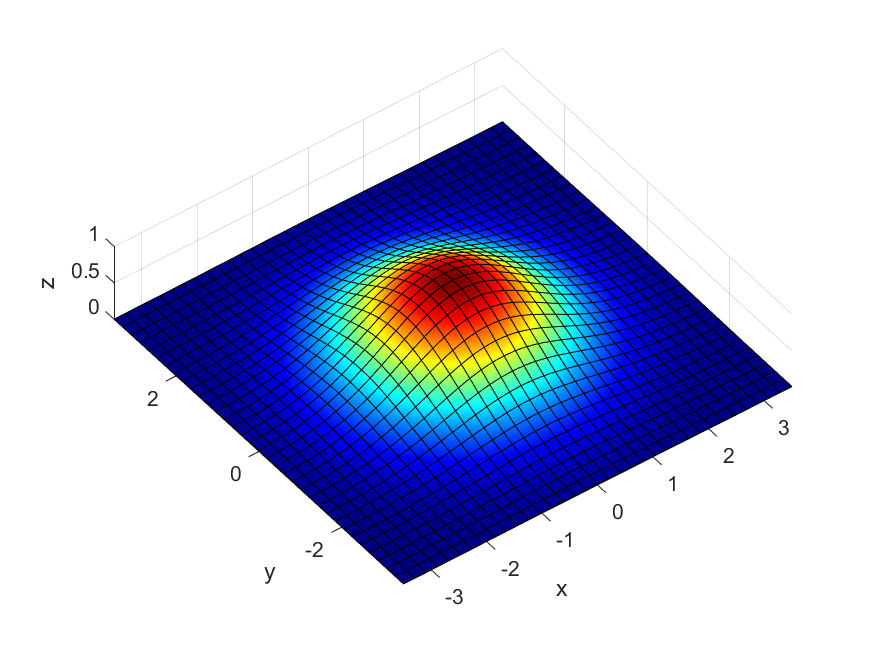} \\
        \hline
        Line & \begin{tabular}{@{}l@{}} Line start $\hat{x}_1$ \\ Line end $\hat{x}_2$ \\ Repulsion matrix $\mathrm{A}$ \end{tabular} & \begin{equation*}
            \begin{matrix}\rho(x) =\frac{(\hat{x}_2-\hat{x}_1)\cdot(x-\hat{x}_1)}{\|\hat{x}_2-\hat{x}_1\|^2}\\
            \\
         \bar{x}_l(x) =x-\hat{x}_1+\mathop{\rm clamp}\left(\rho(x) ,0,1\right)(\hat{x}_2-\hat{x}_1)\end{matrix}
        \end{equation*} & \includegraphics[width=0.09\textwidth,trim={2.5cm 1.5cm 2.25cm 2.5cm},clip]{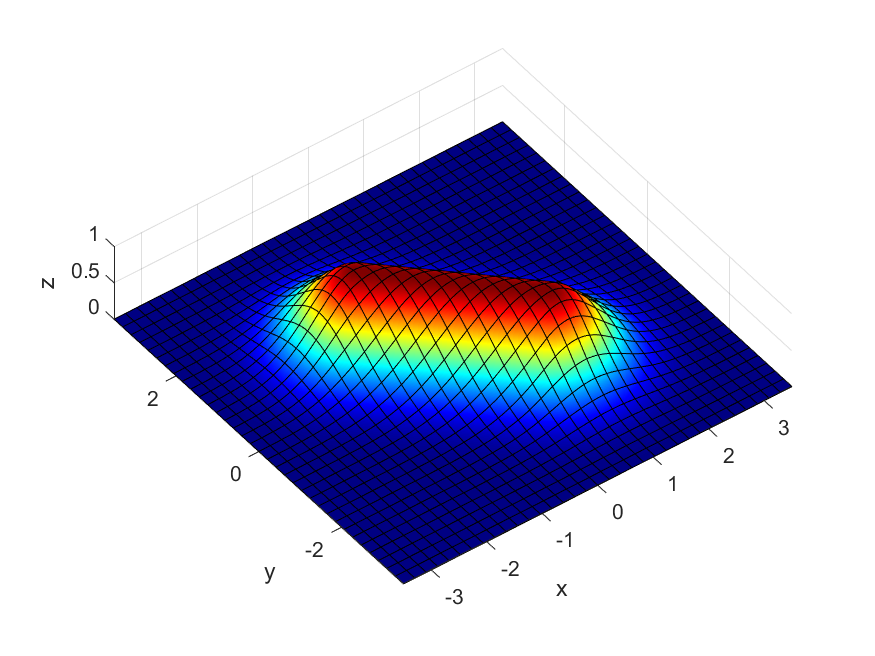} \\
        \hline
        Polygon & 
        \begin{tabular}{@{}l@{}} Vertices $V=\{\hat{x}_1, \dots, \hat{x}_n\}$ \\ Repulsion matrix $\mathrm{A}$ \end{tabular} & 
        \begin{equation*}
            \begin{matrix}
                \bar{v}_i(x) = \text{Line}_{\bar{x}_l}(x, \hat{x}_i, \hat{x}_{i+1}) \\
                k = \mathop{\text{argmin}}\limits_{i} \|\bar{v}_i(x)\|_{\mathrm{A}} \\
                \bar{x}_{po}(x) = \begin{cases} 
                    0 & x \in \mathcal{P} \\ 
                    \bar{v}_k(x) & \text{otherwise} 
                \end{cases}
            \end{matrix}
        \end{equation*}
        & \includegraphics[width=0.09\textwidth,trim={2.5cm 1.5cm 2.25cm 2.5cm},clip]{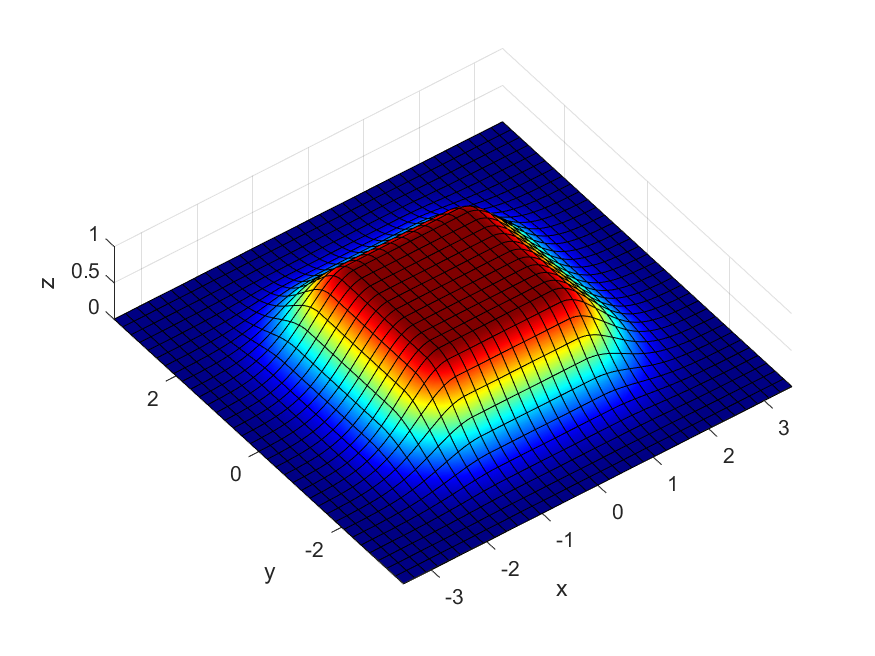} \\
        \hline
        \textit{Collection} & \textit{Parameters of the sub units} & \textit{Repulsion vector $\bar{x}(x)$ of sub unit with smallest Mahalanobis distance $\tilde{d}(x)$} & \includegraphics[width=0.09\textwidth,trim={2.5cm 1.5cm 2.25cm 2.5cm},clip]{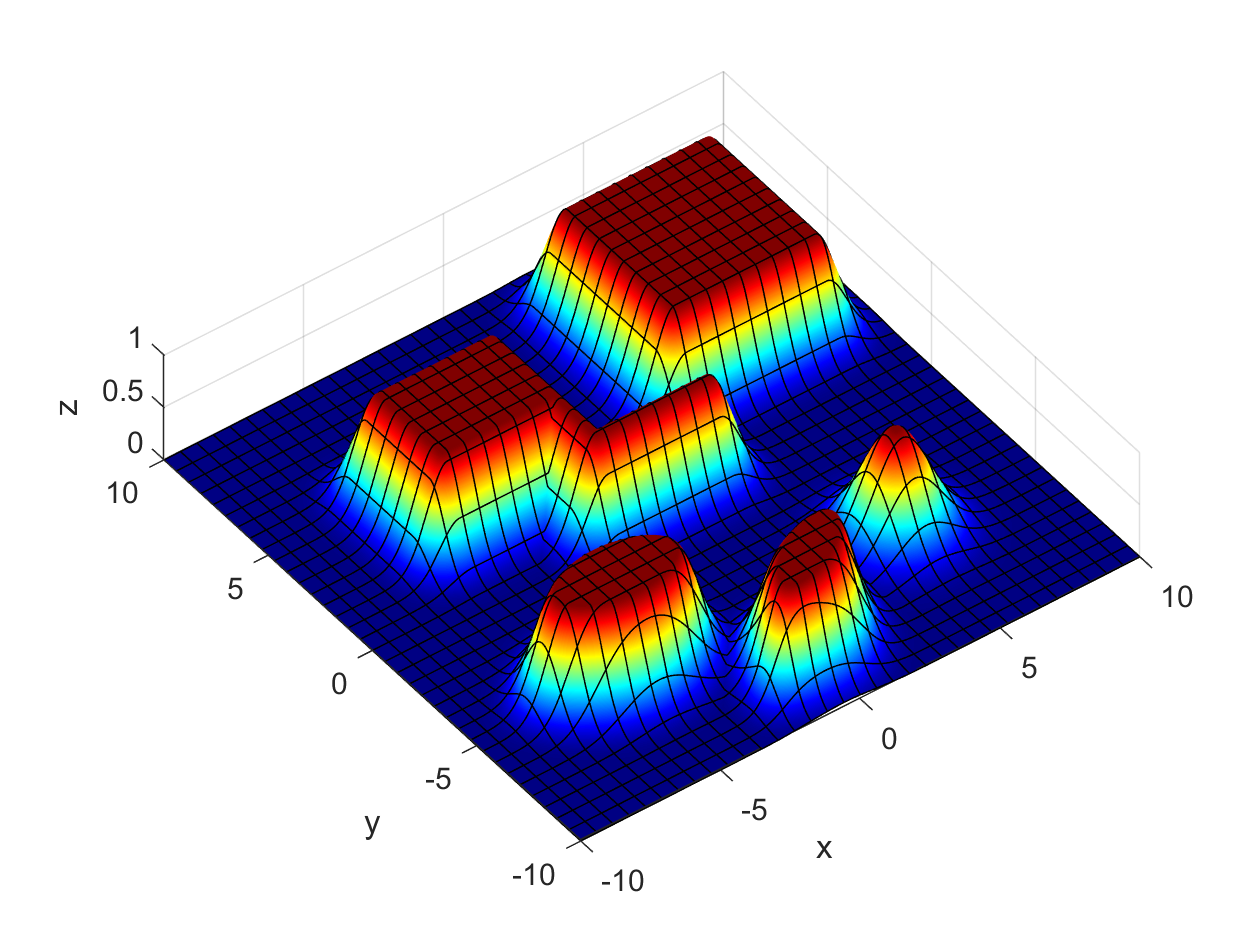} \\
        \hline
    \end{tabular}
\end{table*}

\begin{definition} [Squared Distance to an Unit]
A proxy to the squared distance to a risk field unit $d\,^2(x)$ is defined as: $$d\,^2(x) = \bar{x}(x)^T \bar{x}(x),$$ where $\bar{x}(x)$ is the repulsion vector detailed in Table \ref{tb:fields} and $x$ is an evaluated point. For a collection of units, $d\,^2(x)$ is the smallest squared distance among all sub-units.
\end{definition}

\begin{definition} [Squared Mahalanobis Distance to an Unit]
The squared Mahalanobis distance to a risk field unit $\tilde{d}\,^2(x)$ is defined as: $$\tilde{d}\,^2(x) = \bar{x}(x)^T \mathrm{A}^{-1} \bar{x}(x),$$ where $\bar{x}(x)$ is again the repulsion vector from Table \ref{tb:fields}, $\mathrm{A}$ is the positive definite repulsion matrix of the unit, and $x$ is an evaluated point. In the case of a collection of units, $\tilde{d}\,^2(x)$ is the minimum of the squared Mahalanobis distances with respect to its sub-units.
\end{definition}

\begin{definition} [Risk with Respect to an Unit]
The risk with respect to a unit $\sigma (x)$ is defined as the exponential of the squared Mahalanobis distance to the unit: $$\sigma (x) = \exp{\left(\tilde{d}\,^2(x)\right)},$$ where $\tilde{d}\,^2(x)$ is the squared Mahalanobis distance, and $x$ is an evaluated point.
\end{definition}

%% file: chapters/larp.tex
\section{Methodology}
\label{sec:method}

\begin{figure}[t!]
  \centering
    \begin{subfigure}[b]{0.23\textwidth}
        \centering
        \includegraphics[width=\textwidth]{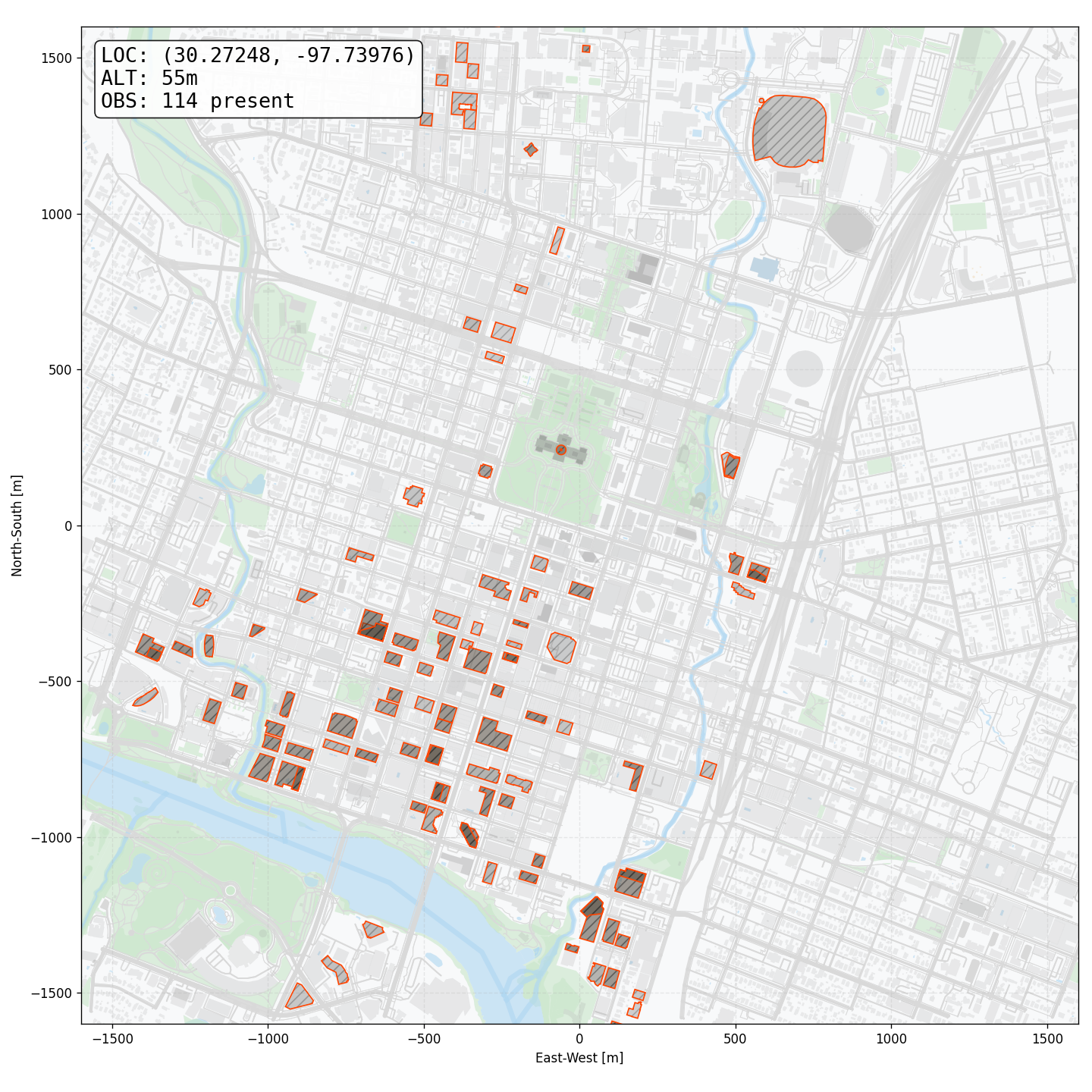}
        \caption{Austin, TX, USA}
    \end{subfigure}~
    \begin{subfigure}[b]{0.26\textwidth}
        \centering
        \includegraphics[width=\textwidth]{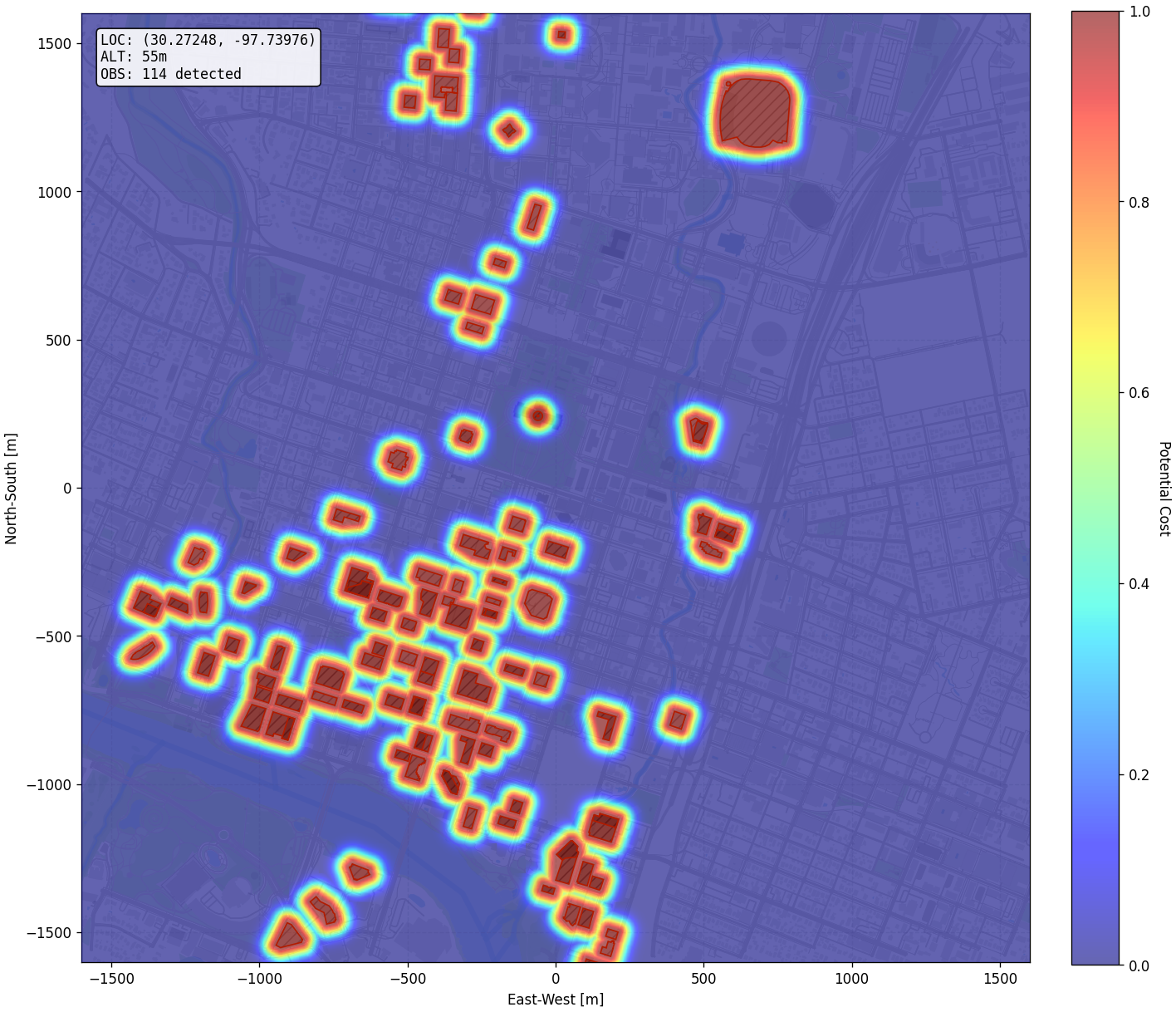}
        \caption{Corresponding Risk Field}
    \end{subfigure}
    \begin{subfigure}[b]{0.38\textwidth}
        \centering
        \includegraphics[width=\textwidth]{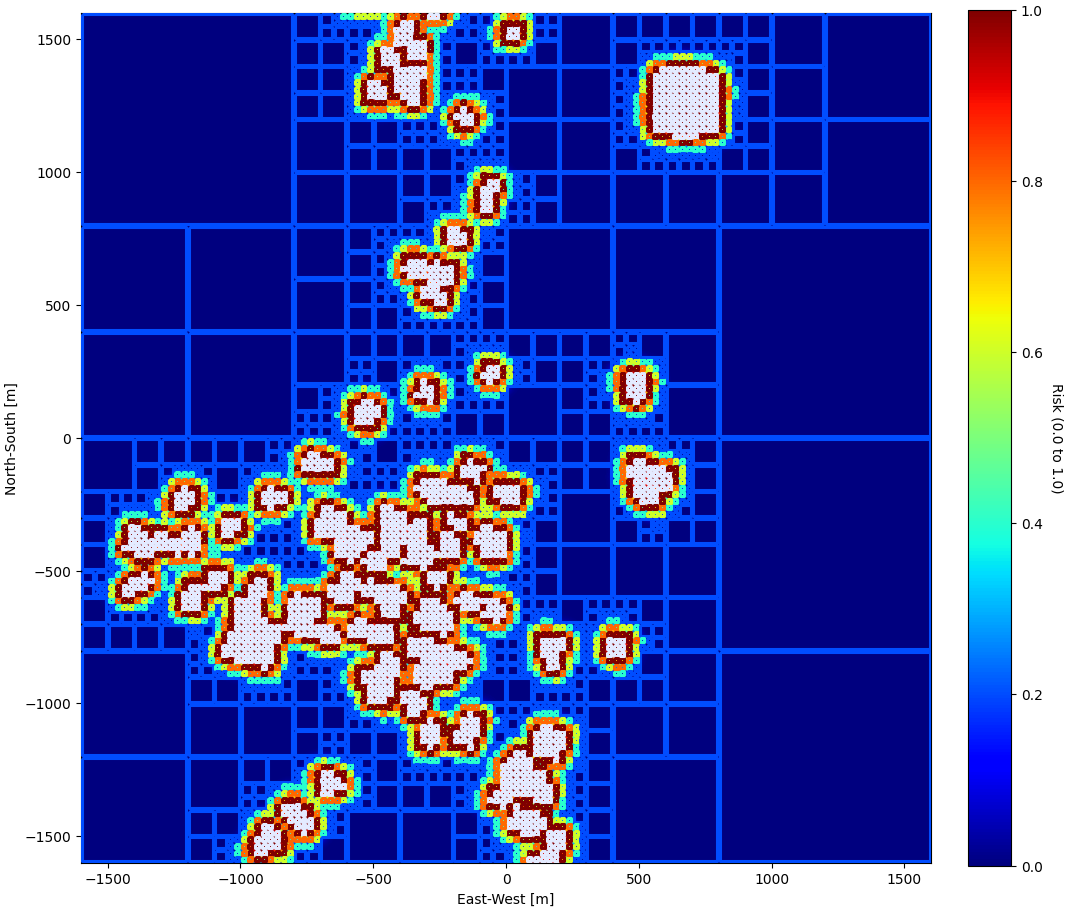}
        \caption{Risk-Aware Multi-Scale Cell Decomposition}
        \label{fig:cell-decomp}
    \end{subfigure}
    \caption{Risk-aware multi-scale cell decomposition of Austin, TX (50–60m AGL). White cells mark obstacle zones; color denotes risk level.}
    \label{fig:austin-map}
\end{figure}

Leveraging the properties of the standardized risk field defined in Section \ref{sec:units}, we introduce \textit{Larp Path Planner} (Last-mile Autonomous Route Planning Path Planner), a framework designed for risk-aware urban air mobility. The core of this framework is a multi-scale decomposition algorithm that partitions an urban environment into distinct risk-aware sectors. These sectors form the foundation for a low-altitude routing graph that prioritizes risk minimization. Fig. \ref{fig:austin-map} demonstrates this decomposition applied to Austin, TX, USA, for an operating altitude of 50--60 meters AGL within a 1.6 km radius.

\subsection{Multi-Scale Risk-Aware Cell Decomposition}

Drawing inspiration from adaptive quad tree structures, the primary phase of our method involves partitioning the risk field into a hierarchy of cells. Each cell is assigned a discrete "Risk Zone" representing the maximum potential for restriction violation within its bounds. The subdivision strategy is adaptive: cells proximal to obstacles are refined to a higher granularity (smaller size), while open airspace is represented by larger cells. This approach significantly reduces the search space while augmenting routing precision near hazards. The decomposition process is detailed in Algorithms \ref{alg:cell-decomp} and \ref{alg:cell-zone}.

\subsubsection{Quad Tree Structure Construction}

Algorithm \ref{alg:cell-decomp} outlines the recursive subdivision of the field. The process begins by initializing a root node representing the entire map. For each quadrant, the maximum risk is estimated via Algorithm \ref{alg:cell-zone}. Based on this estimate, a decision is made to either classify the node as a leaf or subdivide it further. 

\begin{figure}[tb]
  \centering
    \includegraphics[width=0.41\textwidth]{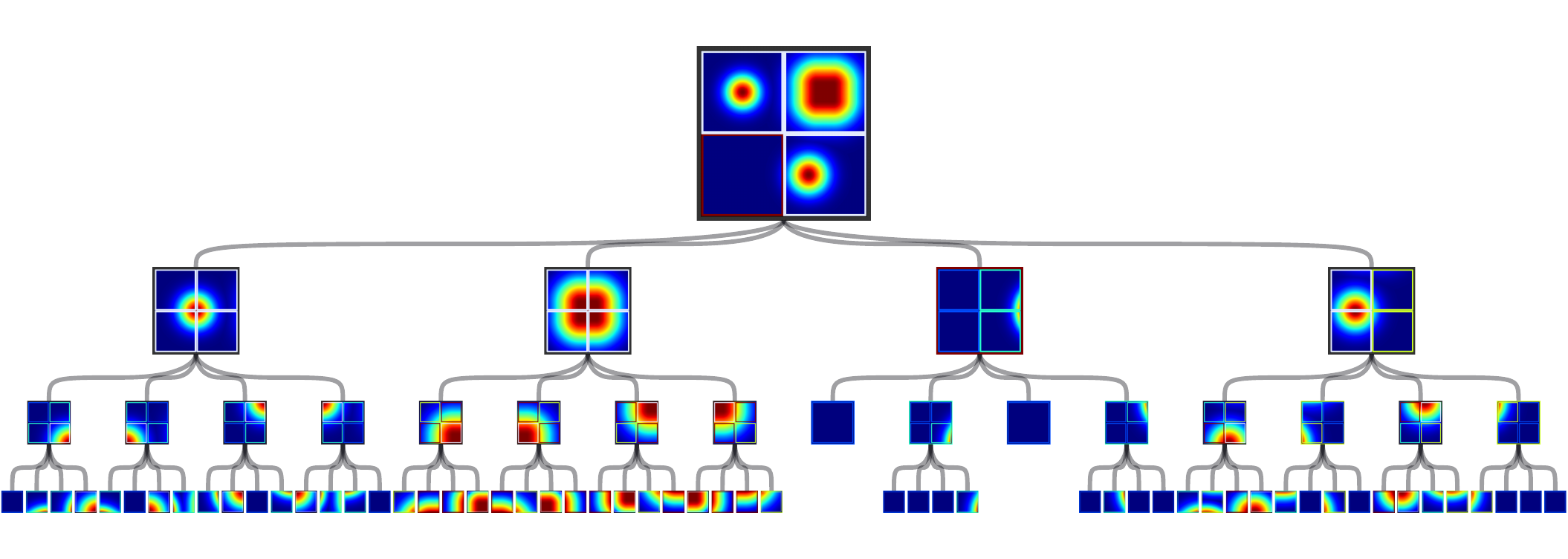}
    \caption{Quad tree subdivision of a simple risk field. Cells refine recursively until risk falls below the safety threshold or minimum resolution is reached.}
    \label{fig:field-tree}
\end{figure}

Subdivision continues until one of two stopping criteria is met: 1) the cell is classified as the safest (lowest risk) zone, or 2) the cell reaches the minimum defined size limit ($n_{min}$). To optimize performance, obstacles located far outside the influence range of a specific quadrant are pruned from the set $U$ during recursive calls. The leaves of the resulting tree constitute the multi-scale discretization of the allowable airspace. An example of the quad tree structure is presented in Fig. \ref{fig:field-tree} for a simple field.

\begin{algorithm}[tb]
\caption{Larp Path Planner - Adaptive Cell Decomposition}
\label{alg:cell-decomp}
\begin{algorithmic}[1]
    \Function{Build}{$x$, $n$, $U$}
        \State $quad \gets QuadNode(x, n)$
        \State $zones \gets$ map with default farthest zone
        \If{$\vert U\vert > 0$}
            \State $zones \gets$ \Call{ApproxObstaclesZones}{$x$, $n$, $U$} 
        \EndIf
        \State $quad.zone \gets \min(zones)$ \Comment{Assign conservative risk}
        \If{$n \leq n_{max}$}
            \If{$n \leq n_{min}$ or $quad.zone$ is safest zone}
                \State Mark $quad$ as leaf
                \State \Return{$quad$}
            \EndIf
        \EndIf
        \State $U \gets \{u \in\, U\, \vert \,zones[u] < \text{safest zone}\}$
        \ForEach{quad $q$ of a cell}
            \State $quad.child[q] \gets$ \Call{Build}{$q.center$, $n/2$, $U$}
        \EndFor
        \State \Return{$quad$}
    \EndFunction
\State $root \gets$ \Call{Build}{\textit{field.center}, \textit{field.size}, \textit{all units}}
\end{algorithmic}
\end{algorithm}

\subsubsection{Risk Estimation and Zone Assignment}

The heart of the decomposition efficiency is the ability to analytically estimate the risk upper bound of a cell without exhaustive sampling. This is achieved via Algorithm \ref{alg:cell-zone}. 

First, we calculate the squared Euclidean distance from the cell center to each obstacle. If this distance is less than the cell's circumscribed radius, the cell potentially contains the obstacle and is immediately assigned to Zone 0 (Obstacle Zone). 

For obstacles outside the cell boundary, we utilize the Mahalanobis distance. Using the obstacle's repulsion vector, we project a point $c$ on the circumscribed circle boundary closest to the obstacle. The risk $\sigma(c)$ is evaluated at this point. The maximum risk contribution from any obstacle determines the cell's final Zone assignment. This guarantees that the assigned Zone represents a conservative upper bound of the risk within that cell.

\begin{algorithm}[tb]
\caption{Larp Path Planner -- Zone Risk Estimation}
\label{alg:cell-zone}
\begin{algorithmic}[1] \Require {Sorted list of boundaries $bds$ in descending order}
\Function{ApproxObstaclesZones}{$x$, $n$, $U$}
    \State $zones \gets$ default safe zone map 
    \ForEach{unit $u$ in $U$}
        \If{$d_u^2(x) \leq \frac{n^2}{2}$}
            \State $zones[u] \gets 0$  \Comment{Obstacle inside cell}
        \EndIf
    \EndFor
    \ForEach{unit $u$ in $U$ not in $zones$}
        \State $c \gets x - \frac{n}{\sqrt{2}\,\|\bar{x}_u(x)\|}\, \bar{x}_u(x)$ \Comment{Project to edge}
        \State $p \gets \exp{\left(\Tilde{d}_u^2(c)\right)}$
        \State $zones[u] \gets \text{Bin}(p, bds)$
    \EndFor
    \State \Return{$zones$}
\EndFunction
\end{algorithmic}
\end{algorithm}

\begin{figure}[tb]
  \centering
    \includegraphics[width=0.25\textwidth]{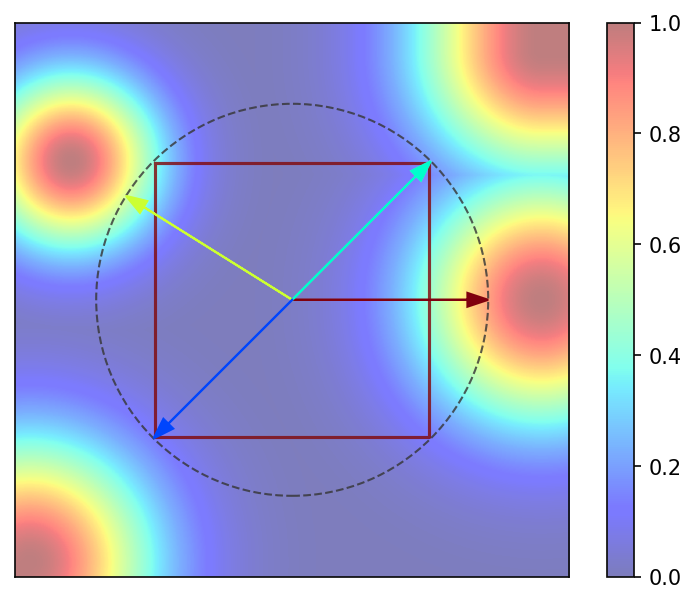}
    \caption{Conservative risk upper bound estimation for a cell. Risk is evaluated at the intersection of the cell's circumscribed circle and the obstacle's repulsion vector.}
    \label{fig:field-area}
\end{figure}

Fig. \ref{fig:field-area} illustrates this approximation. The repulsion vectors are projected from the center to the extent of the cell's diagonal. The magnitude of the vector and the risk at the boundary inform the maximum possible risk $\sigma_{max} \approx 0.8$ for that region, allowing for rapid classification without dense grid sampling.

\subsection{Risk-Aware Global Planning}

\subsubsection{Network Graph Construction}

Following decomposition, a connectivity graph is constructed. A divide-and-conquer strategy traverses the quad tree to identify adjacency relationships between leaf nodes. This transforms the continuous risk field into a discrete weighted graph where nodes represent risk-aware sectors and edges represent feasible transitions.

\subsubsection{Path Search}

To generate optimal routes, we employ a modified A* algorithm. The standard Euclidean distance cost function is augmented to penalize high-risk traversals:
\begin{equation}
    d(q_{a}, q_{b}) = s(q_{b}) \cdot \| q_a.\text{center} - q_b.\text{center}\|
\end{equation}
where $q_a$ and $q_b$ denote the current and target quadrants, respectively, and $s(q_b) \geq 1.0$ is a scaling factor derived from the Risk Zone of $q_b$. This effectively dilates the distance of high-risk edges, guiding the solver toward safer corridors unless a significant shortcut justifies the increased risk.

\begin{figure}[tb]
  \centering
  \includegraphics[width=0.30\textwidth]{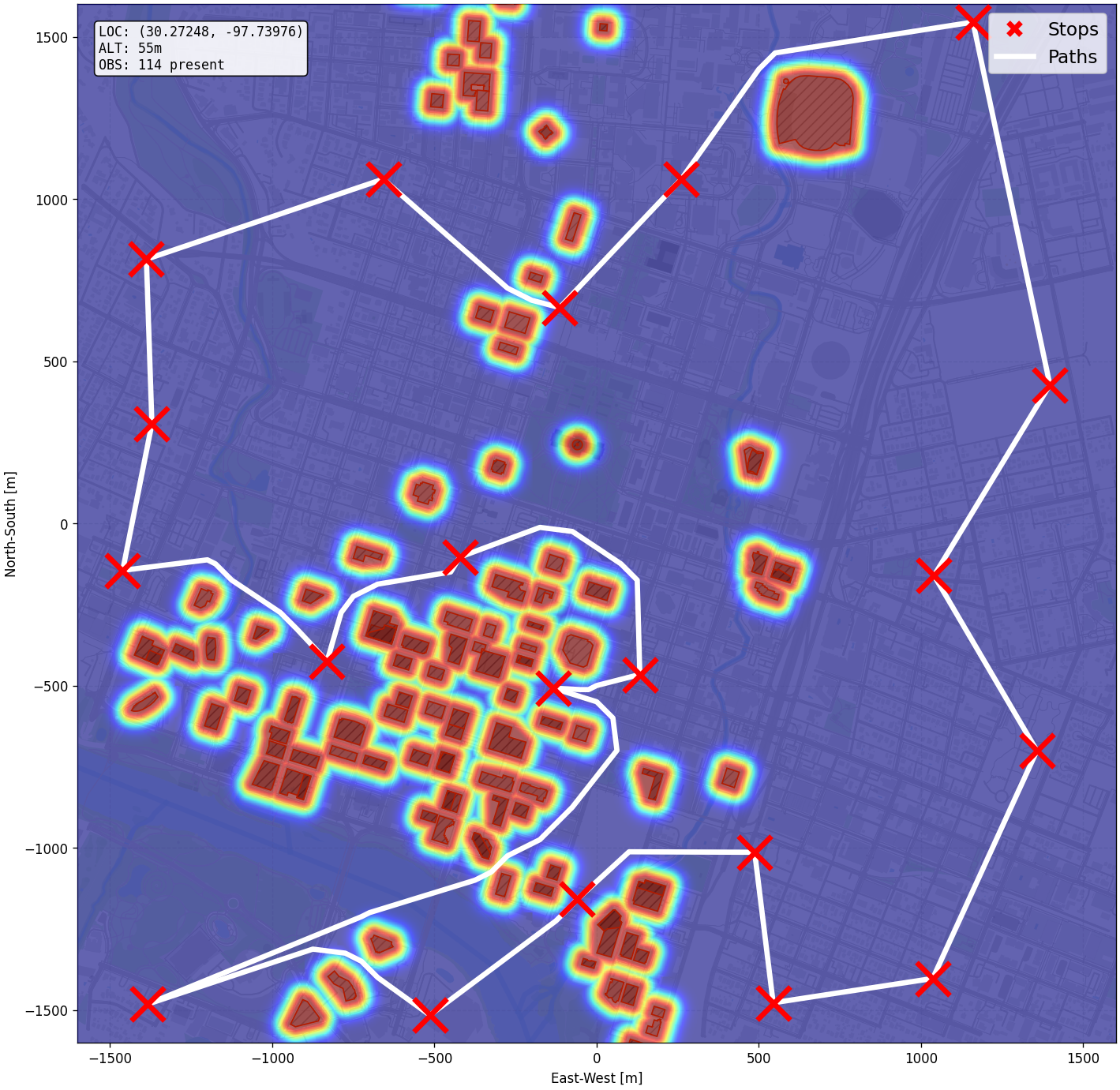}
  \caption{Aggregated delivery routes over Austin, TX (55m AGL), demonstrating efficient traversal of open areas and dense downtown corridors.}
  \label{gr:autin-routes}
\end{figure}

Once a sequence of nodes is identified, the path is smoothed using line-of-sight optimization to produce flight-ready waypoints. Fig. \ref{gr:autin-routes} depicts the resulting aggregated delivery routes for the Austin simulation.

%% file: chapters/results.tex
\section{Experiments and Results}
\label{sec:experiments}

\subsection{Experimental Setup}

The simulations utilized real-world urban data extracted via OpenStreetMap (OSM) for five distinct urban environments, chosen to represent varying degrees of obstacle density complexity. The search space for each environment was defined by a specific operating altitude and a radial extent ranging from 1.0 km to 2.5 km:

\begin{enumerate}
    \item \textbf{Austin, TX (Downtown) \& Boston, MA (Seaport):} Mid-rise urban layouts with irregular boundaries (Altitude: 50m; Radius: 1.6 km).
    \item \textbf{Singapore (Marina Bay \& Lucky Plaza):} A mixture of open water bodies and ultra-dense commercial corridors (Altitude: 50--70m; Radius: 1.0--2.0 km).
    \item \textbf{Hong Kong (Central):} An environment characterized by extreme verticality and "urban canyons" (Altitude: 150m; Radius: 2.5 km).
\end{enumerate}

We benchmark our \textit{Adaptive Larp Path Planner} against three baselines: 
1) \textbf{M-APF} \cite{rostami2019obstacle, bounini2017modified}, a local reactive planner with anisotropic repulsion; 
2) \textbf{Informed RRT*} \cite{gammell2014informed, gammell2015batch}, a sampling-based asymptotically optimal planner (averaged over multiple stochastic trials); and 
3) \textbf{Larp Path Planner (Fixed Grid)}, an ablation study using a uniform grid decomposition to isolate the benefits of the multi-scale approach.

Performance metrics include Success Rate (percentage of collision-free paths), Compute Time, Peak Risk (maximum risk encountered), Cumulative Risk (path integral of risk), and Path Length.

\subsection{Comparative Analysis}

Table \ref{tb:city-comparison} summarizes the aggregate performance across 750+ path planning queries.

\subsubsection{Success Rate and Efficiency}
The proposed Multi-Scale Larp Path Planner achieved a 100\% success rate across all tested environments. In contrast, local reactive methods (M-APF) suffered in dense environments like Lucky Plaza and Hong Kong, frequently becoming trapped in local minima (73.5\% success rate).

In terms of computational cost, our method demonstrated order-of-magnitude improvements over sampling-based approaches. While Informed RRT* required an average of $6.42 \pm 6.97$ seconds to converge to a solution, Larp Path Planner (Adaptive) generated paths in $0.01 \pm 0.01$ seconds. This rapid computation is critical for enabling real-time replanning in dynamic UAM operations.

\subsubsection{Risk Minimization}

A key contribution of this work is the explicit minimization of risk. As shown in Table \ref{tb:city-comparison}, Larp Path Planner yielded the lowest Cumulative Risk ($12.04$). While Informed RRT* produces short paths, it often "clips" corners of high-risk zones to optimize for Euclidean distance, resulting in significantly higher cumulative risk ($38.43$) and peak risk ($0.24$). Our method balances path length with risk avoidance, resulting in slightly longer paths (approx. 7\% increase over RRT*) that are significantly safer.

\begin{table*}[t]
    \caption{Quantitative Performance Comparison Across 750+ Path Planning Trials in Five Urban Environments.}
    \label{tb:city-comparison}
    \centering
    \begin{tabularx}{\textwidth}{@{}|l|C C C C C|@{}}
        \toprule
        \textbf{Algorithm} & \textbf{Success Rate} & \textbf{Compute Time [s]} & \textbf{Peak Risk [$\sigma$]} & \textbf{Cumul. Risk [$\sigma$]} & \textbf{Path Length [m]}\\
        \midrule
        M-APF & 73.5\% & 0.21 $\pm$ 0.28 & \textbf{0.04 $\pm$ 0.07} & 13.56 $\pm$ 32.87 & \textbf{701.99 $\pm$ 431.12} \\
        Informed RRT* & 79.0\% & 6.42 $\pm$ 6.97 & 0.24 $\pm$ 0.32 & 38.43 $\pm$ 59.52 & 717.71 $\pm$ 415.24 \\
        Larp Path Planner (Fixed Grid) & \textbf{100.0\%} & 0.07 $\pm$ 0.10 & 0.10 $\pm$ 0.15 & 16.66 $\pm$ 29.65 & 749.21 $\pm$ 511.35 \\
        \textbf{Larp Path Planner (Adaptive)} & \textbf{100.0\%} & \textbf{0.01 $\pm$ 0.01} & 0.08 $\pm$ 0.14 & \textbf{12.04 $\pm$ 24.14} & 769.72 $\pm$ 530.45 \\
        \bottomrule\addlinespace
        \multicolumn{5}{l}{\footnotesize \^~Metrics computed only on successful trials.}
    \end{tabularx}
\end{table*}

\subsection{Scalability: Fixed Grid vs. Multi-Scale Decomposition}

To isolate the benefits of the multi-scale approach, we compared our Adaptive method against a Fixed Grid implementation of the same risk field (Table \ref{tb:city-comparison}, Rows 3 vs 4). 

The Multi-Scale approach reduced computation time by approximately 85\% compared to the Fixed Grid ($0.01$s vs $0.07$s). In large open areas (e.g., over the water in Marina Bay or above parks in Austin), the adaptive quad-tree consolidates space into fewer, larger nodes, drastically reducing the search space for the A* planner. Conversely, in dense clusters, the grid refines automatically to capture narrow corridors. The Fixed Grid approach, forced to maintain high resolution everywhere to ensure safety in narrow passages, suffers from combinatorial explosion in the graph size. This comparison highlights the criticality of multi-scale decomposition for achieving city-scale scalability.

\subsection{Qualitative Results in Dense Urban Sectors}

\begin{figure}[tb]
    \centering
    \subcaptionbox
      {Boston, MA}
      {\bfseries \includegraphics[width=0.23\textwidth]{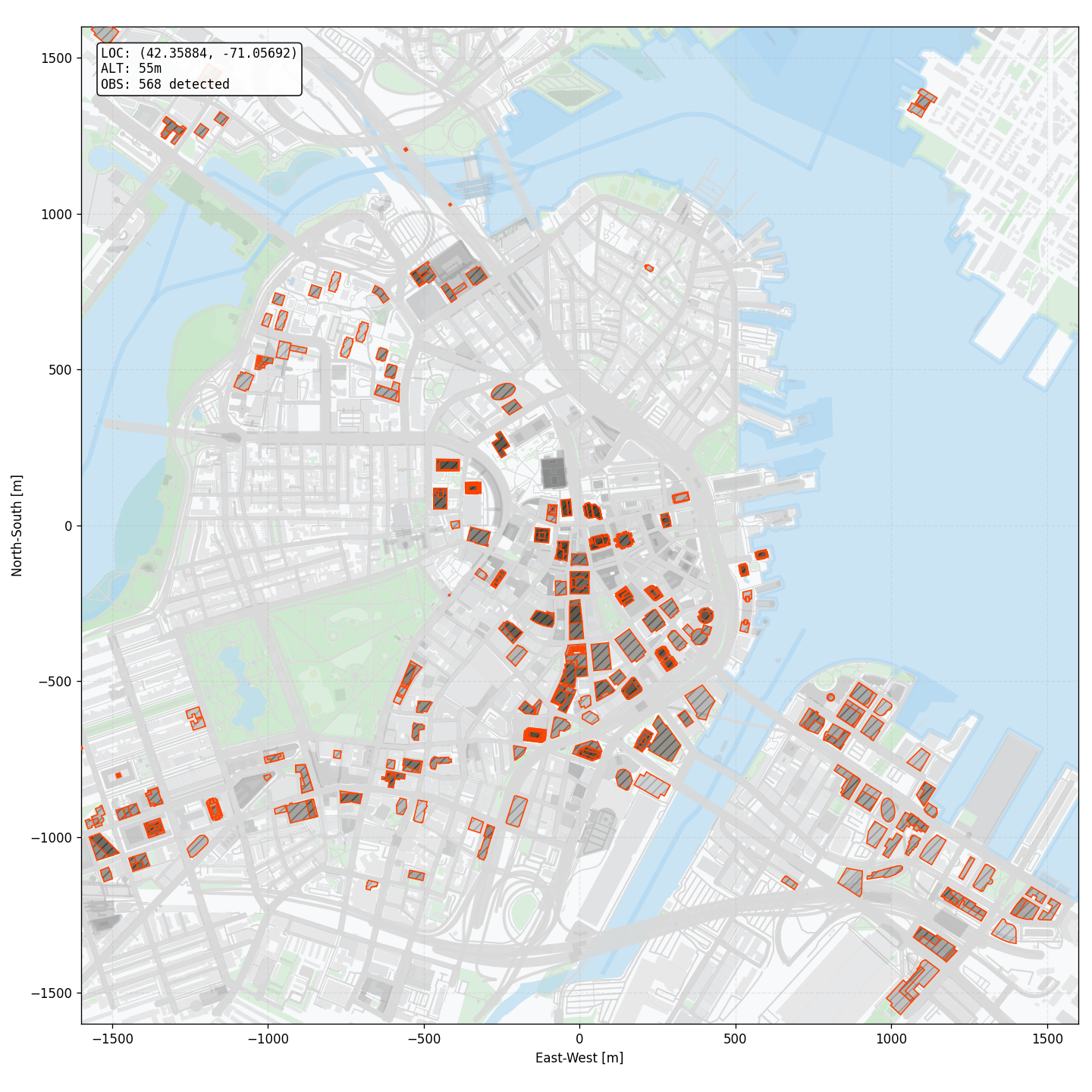}}%
    \subcaptionbox
      {Generated Routes.}
      {\bfseries \includegraphics[width=0.23\textwidth]{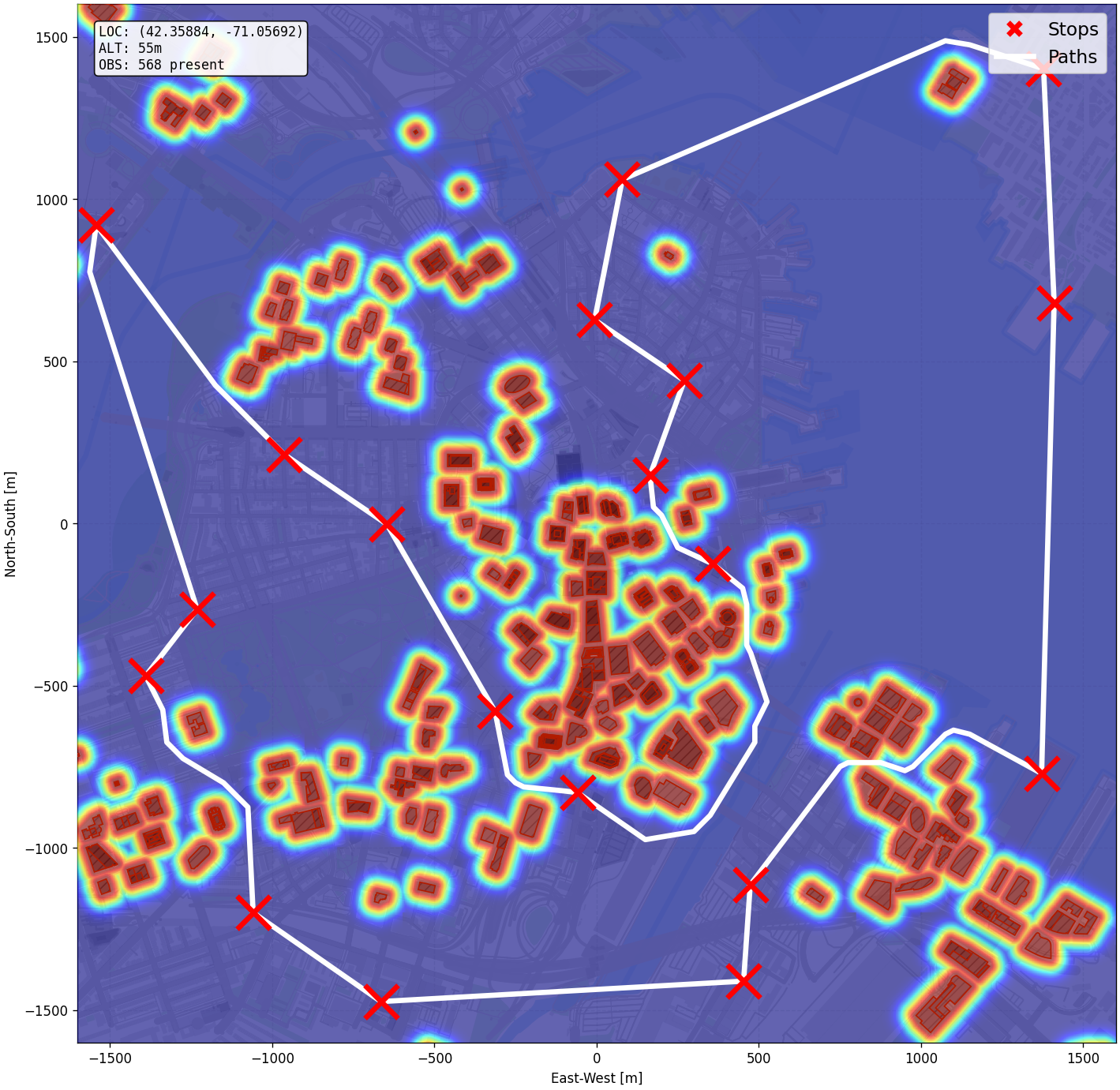}}%
  \caption{Risk field and planned routes for Boston, MA. Safe regions (dark blue) are preferred; high-risk zones (dark red) are avoided.}
    \label{fig:boston}
\end{figure}

Fig. \ref{gr:autin-routes} and \ref{fig:boston} illustrate the generated paths for Boston and Austin, respectively. The risk-aware nature of the planner is evident in the trajectory behavior: the UAV maintains a high clearance from obstacles in open space (minimizing risk) but is capable of traversing narrow "urban canyons" when necessary to reach a destination.

\begin{figure}[tb]
    \centering
    \includegraphics[width=0.40\textwidth]{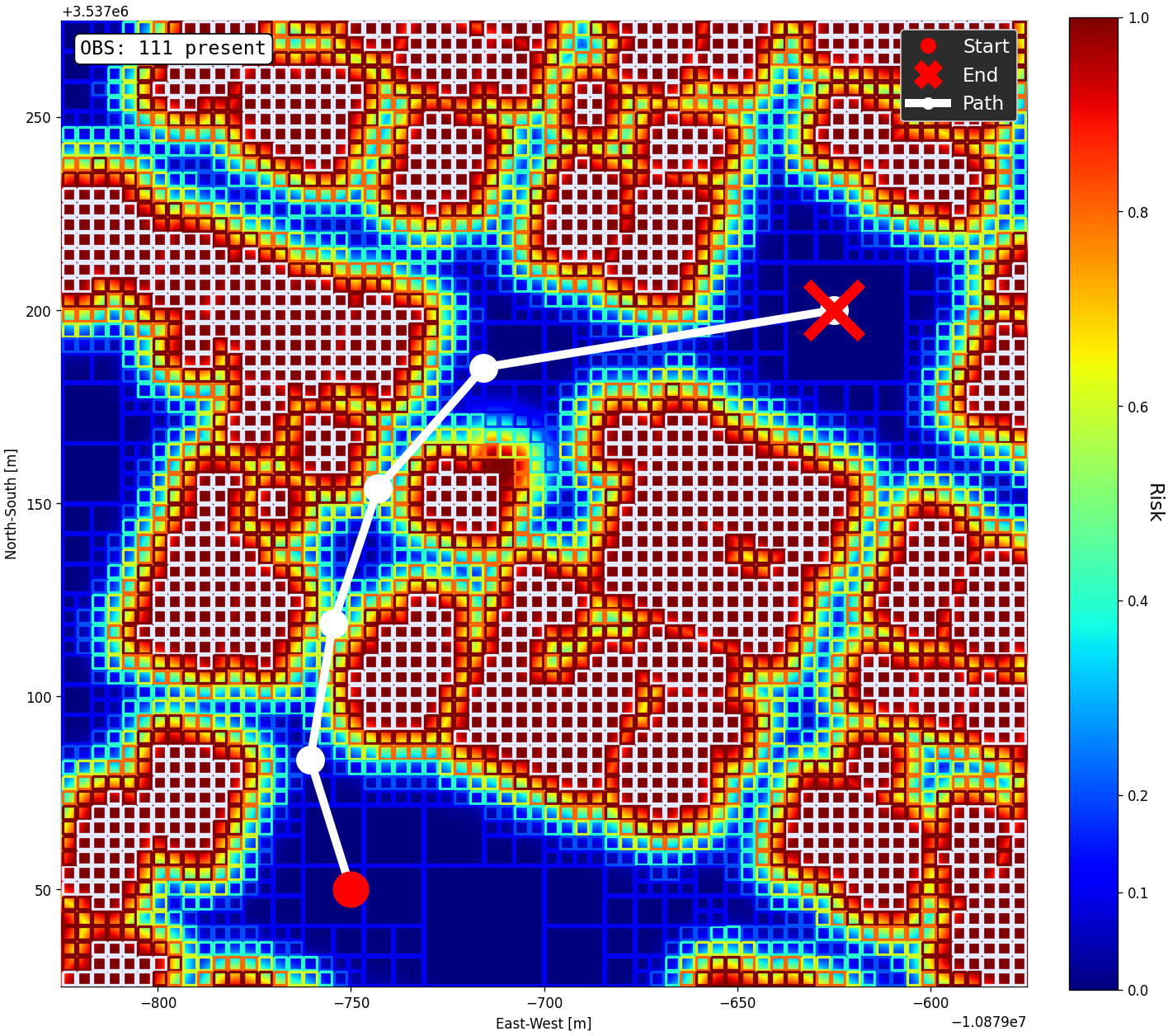}
  \caption{Close-up of Larp Path Planner navigating a dense urban cluster, identifying safe corridors through the building infrastructure.}
    \label{fig:larp-neigh}
\end{figure}

In extremely dense scenarios (Fig. \ref{fig:larp-neigh}), the planner successfully navigates the crevices formed by building clusters. The risk field creates a cost topology that guides the planner through the centerline of avenues, treating the proximity of building surfaces as soft constraints that are only approached when no topological alternative exists.

%% file: chapters/conclusion.tex
\section{Conclusion}
\label{sec:conclusion}

Safe and scalable path planning is a prerequisite for low-altitude UAM, yet 
existing methods fail to simultaneously satisfy computational, safety, and 
city-scale demands. The Larp framework addresses this through risk-aware 
multi-scale cell decomposition, enabling verifiably safe global 
path planning from any GeoJSON map input. Across 750+ trials in five urban 
environments, the adaptive planner achieved a 100\% success rate, the lowest 
cumulative risk of any evaluated method (12.04, a 69\% reduction over Informed 
RRT* (38.43)), and near-real-time computation ($0.01 \pm 0.01$\,s, over 
600$\times$ faster than Informed RRT* and 85\% faster than the fixed-grid 
ablation), with only a $\sim$7\% path length overhead, confirming that 
multi-scale decomposition is critical for city-scale scalability.

Future work will pursue tighter risk bounding to reclaim usable airspace in 
highly restrictive sectors, and extending the static quadtree into a dynamic 
structure for real-time updates, enabling the framework to adapt to moving obstacles and evolving flight restrictions within an active UAM and UTM ecosystem. The source code is 
publicly available at \url{https://github.com/wzjoriv/Larp}.